\documentclass[11pt]{article}

\usepackage[preprint]{acl}

\usepackage{fontspec}
\usepackage{fancyhdr}
\newfontfamily\devanagarifont{Shobhika}[
  Script      = Devanagari,
  Extension   = .otf,
  UprightFont = *-Regular,
  BoldFont    = *-Bold]

\newcommand{\dv}[1]{{\devanagarifont #1}}

\usepackage{amsmath}
\usepackage{amssymb}
\usepackage{graphicx}
\usepackage{microtype}
\usepackage{booktabs}
\usepackage{multirow}
\usepackage{array}
\usepackage{tabularx}
\usepackage{makecell}
\usepackage{float}
\usepackage{xcolor}
\usepackage[most]{tcolorbox}

\newcommand{\system}{\textsc{LipiMix}}
\newcommand{\byt}{ByT5}
\newcommand{\mfive}{mT5}
\newcommand{\cer}{\textsc{CER}}
\newcommand{\wer}{\textsc{WER}}

\definecolor{rawred}{RGB}{220,50,50}
\definecolor{fixgreen}{RGB}{0,150,80}
\definecolor{prepblue}{RGB}{30,100,200}

\definecolor{exframe}{RGB}{31,78,121}    
\definecolor{exback}{RGB}{238,244,250}   
\newtcolorbox{examplebox}{enhanced, breakable=false,
  colback=exback, colframe=exframe,
  boxrule=0.4pt, leftrule=2.5pt, arc=1.5pt, outer arc=1.5pt,
  left=5pt, right=5pt, top=3pt, bottom=3pt,
  before skip=6pt, after skip=6pt,
  fontupper=\small}
\newcommand{\exlabel}[1]{{\color{exframe}\textbf{#1}}}

\hypersetup{
  pdftitle={LipiMix: Transliteration of Code-Mixed Romanized Nepali-English Text to Devanagari Script},
  pdfauthor={Saksham Sapkota, Sagar Giri, Nabin Ranabhat, Aman Shakya},
  pdfkeywords={transliteration, code-mixing, Romanized Nepali, Devanagari, ByT5, mT5, low-resource NLP}
}

\title{\system{}: Transliteration of Code-Mixed Romanized\\
       Nepali--English Text to Devanagari Script}

\author{
  {\hypersetup{linkcolor=black}Saksham Sapkota\thanks{Equal contribution.}} \quad Sagar Giri\footnotemark[1] \quad Nabin Ranabhat \quad Aman Shakya \\[4pt]
  \normalfont\normalsize Department of Electronics and Computer Engineering \\
  \normalfont\normalsize Pulchowk Campus, Institute of Engineering, Tribhuvan University \\
  \normalfont\normalsize Lalitpur, Nepal \\[3pt]
  \normalfont\normalsize\texttt{\{078bct070.saksham,\,078bct071.sagar,\,078bct050.nabin,\,aman.shakya\}@pcampus.edu.np}
}

\fancypagestyle{plain}{%
  \fancyhf{}%
  \fancyfoot[C]{\small\thepage}}
\fancypagestyle{firstpage}{%
  \fancyhf{}%
  \fancyfoot[C]{\footnotesize\itshape Accepted at the Fifth International Conference on Speech and Language Technologies for Low-Resource Languages}}
\begin{document}
\maketitle
\thispagestyle{firstpage}

\begin{abstract}
Informal Nepali digital communication heavily relies on \emph{code-mixed} text, interleaving Romanized Nepali with English within the same utterance. These highly inconsistent and non-standardized writing styles remain challenging for NLP systems trained on formal Devanagari corpora. We present \textbf{\system{}}, a low-resource system for transliterating code-mixed Romanized Nepali-English text into Devanagari script. Our contributions are threefold. First, we build a large-scale corpus by collecting 2.77M+ YouTube and Facebook comments, retaining 444K+ usable Romanized sentences after cleaning, and constructing a \emph{Phonetic Variant Dictionary} (PVD) that maps 92.7K+ spelling variants to 15.6K+ canonical forms. Second, we create a manually annotated parallel corpus of 20.5K sentence pairs, expanded to 85.8K+ via PVD-based augmentation. Third, we fine-tune \byt{} and \mfive{} seq2seq models on this corpus. Our best model, \byt{} fine-tuned on augmented data, achieves \cer{} of 5.96\% and \wer{} of 21.85\%, outperforming several publicly available Nepali transliteration systems. Homophonic ambiguity and diacritic errors remain key limitations. All the datasets, model checkpoints, and code are publicly released at \url{https://github.com/oceangiri23/Transliteration}.
\end{abstract}

\section{Introduction}
\label{sec:intro}
 
Nepali, spoken by approximately 17 million native speakers \cite{eberhard2023ethnologue}, has a large and active social-media presence where informal communication routinely exhibits \emph{code-mixing} -- the fluent alternation between Romanized Nepali and English within a single utterance \cite{dogruoz2021survey,winata2023codeswitching,pandey2025codemixed} -- as in the naturally occurring comment:

\begin{examplebox}
\exlabel{Romanized Nepali + English:}\\
\emph{``Ekdam few manxe haru matrw extra hunxan, aru sab normal life nai bachirako hunxa ni''}\\[0.3em]
\exlabel{English Translation:}\\
\emph{``Only a very few people are extra, everyone else is just living a normal life''}
\end{examplebox}

Yet Devanagari remains the target format for most downstream NLP tasks (sentiment analysis \cite{chaudhary2025neptun}, hate-speech detection, machine translation \cite{kartik2024synthetic}, information retrieval), leaving such Romanized code-mixed content poorly supported.
 
\paragraph{Challenges.}
Transliterating Romanized Nepali is hard for two reasons. \emph{Orthographic inconsistency} is extreme: e.g., the word for \emph{it will happen} appears as \emph{huncha}, \emph{hunchha}, \emph{hunxa}, \emph{hunachha}, among other variants. \emph{Homophonic ambiguity} is also prevalent: \emph{kala} may mean \dv{कला} (art) or \dv{काला} (black); \emph{bal}, \dv{बल} (strength) or \dv{बाल} (hair) - requiring sentence-level, not word-level, disambiguation. These challenges are compounded by an acute scarcity of annotated parallel resources: unlike Hindi-English \cite{srivastava2020phinc} or Spanish-English \cite{dogruoz2021survey}, Nepali-English code-mixed NLP remains severely under-resourced \cite{pandey2025codemixed,kafley2024transliteration}.
 
\paragraph{Limitations of existing approaches.}
Rule-based mappers \cite{asheshunicode,kokilunicode,nepaliunicodecom,bewakes_unicode,easynepalityping}, and Google Transliteration \cite{googletranslit} target only formally Romanized text and struggle with colloquial code-mixed input. NepaliXlit~\cite{patel2026from}, the leading open model, fine-tunes IndicXlit \cite{madhani2023aksharantar} but is likewise trained on formal Romanized text and struggles with colloquial variants (e.g., \emph{manchey} for \dv{मान्छे}). Word-level datasets \cite{kafley2024transliteration} and normalization studies \cite{chaudhary2025neptun} exist, but no publicly available sentence-level transliteration system for code-mixed Nepali-English currently exists.
 
\paragraph{Contributions.}
Our models build on standard sequence-to-sequence architectures; the primary contribution is instead the \emph{data engineering pipeline} that makes them effective for noisy, code-mixed Romanized Nepali:
\begin{enumerate}\setlength\itemsep{3pt}
  \item \textbf{Large-scale corpus.} We collect 2,772,519 Nepali
        YouTube and Facebook comments and release 444,116 cleaned Romanized
        sentences after a three-stage filtering pipeline.
  \item \textbf{Phonetic Variant Dictionary (PVD).} To address extreme spelling variability, we construct a Phonetic Variant Dictionary mapping 92,762 spelling variants to
        15,651 canonical base forms, enabling both inference-time
        normalization and data augmentation.
  \item \textbf{Parallel corpus.} We manually annotate 20,500
        Romanized Nepali-English / Devanagari sentence pairs, split into 19.5k/1k train and validation splits respectively, and the training data was
        expanded to 85,870 via PVD-based augmentation.
  \item \textbf{Fine-tuned models.} We fine-tune \byt{} and \mfive{}
        on both original and augmented corpus with the metrics shown in Section~\ref{ssec:quant}; the best model (\byt{} Augmented) achieves \cer{}~=~5.96\% and \wer{}~=~21.85\% on our separately annotated raw social media test set outperforming existing publicly available Nepali transliteration systems, as shown in Section~\ref{ssec:baselines}.
  \item \textbf{Public and reproducible release.} All the datasets, model checkpoints, pre-processing scripts, PVD construction code, and training configurations are publicly available under a permissive license.
\end{enumerate}
 
\section{Related Work}
\label{sec:related}

\paragraph{Romanized Nepali processing.}
Early work on Romanized Nepali was primarily rule-based. \citet{bewakes_unicode} proposed a character-level mapper for formal Romanized Nepali, while \citet{kafley2024transliteration} introduced a word-level transliteration dataset, limiting sentence-level applications. \citet{chaudhary2025neptun} studied normalization for sentiment analysis, and \citet{bhat2014normalization} addressed transliterated search for Indic languages that motivate our approach to code-mixed Nepali.

\paragraph{Code-mixed NLP.}
Code-mixed NLP has been studied extensively across language pairs \cite{dogruoz2021survey,winata2023codeswitching,das2014identifying,pratapa2018language}, particularly Hindi--English \cite{srivastava2020phinc,mave2018language}. Synthetic data and augmentation have shown benefits for low-resource code-mixed translation and pretraining \cite{kartik2024synthetic,hamed2023augmentation,wang2025codeswitching}, motivating our augmentation strategy. Word-level language identification methods \cite{rijhwani2017estimating} are also relevant to our English-token filtering.

\paragraph{Indic transliteration resources.}
Dakshina \cite{roark2020dakshina} and Aksharantar \cite{madhani2023aksharantar} provide large-scale transliteration data for Indic languages, while IndicXlit \cite{madhani2023aksharantar} leverages cross-lingual transfer across 21 languages. NepaliXlit \cite{patel2026from} adapts IndicXlit to Nepali but remains less robust on colloquial Romanized text (Section~\ref{ssec:baselines}). Prior work also highlights challenges in romanization and back-transliteration for low-resource languages \cite{amrhein2020romanization,gella2013query}.

\paragraph{Limitations of existing approaches.}
Existing resources do not directly address noisy, code-mixed Romanized Nepali--English. Hindi--English resources \cite{srivastava2020phinc,mave2018language} generally assume more standardized Romanization, while Aksharantar/IndicXlit \cite{madhani2023aksharantar,kunchukuttan2021largescale} focus on formally Romanized word pairs. Synthetic augmentation methods \cite{kartik2024synthetic,hamed2023augmentation,wang2025codeswitching} target translation or language modeling rather than orthographic normalization. These limitations motivate our Phonetic Variant Dictionary and augmentation pipeline for noisy Nepali social-media text.

\paragraph{Sequence-to-sequence models for transliteration.}
Neural seq2seq models are widely used for transliteration \cite{vaswani2017attention,raffel2020t5}. Byte-level models \cite{xue2022byt5} are robust to spelling variation \cite{clark2022canine,yu2023megabyte}, while multilingual pretrained models \cite{xue2021mt5,conneau2020unsupervised} provide cross-lingual priors but may poorly tokenize irregular Romanized spellings. We study this trade-off in Section~\ref{sec:results}. Unlike prior work, we focus on sentence-level transliteration of informal, code-mixed Romanized Nepali--English text, combining orthographic normalization with code-switch preservation.

\section{Data Collection and Corpus Construction}
\label{sec:dataset}

\subsection{Collection and Cleaning}
\label{ssec:collection}

We collected \textbf{2,772,519} comments from 10 Nepali YouTube channels and public Facebook pages across four genres: News \& Politics, Music \& Entertainment, Infotainment, and Tech \& Finance (per-source counts in Appendix~\ref{sec:data_collection}, Table~\ref{tab:raw_comments}). A three-stage pipeline comprising \emph{preliminary cleaning}, \emph{Romanization filtering}, and \emph{quality filtering} removed non-textual content, pure Devanagari comments, comments exceeding the English-token threshold ($\tau=0.6$), and comments with insufficient genuine Romanized Nepali content. This reduced the corpus to 444K final sentences. A manual spot-check of 200 comments found 96\% well-formed, supporting the generalization of the filtering pipeline beyond the annotated subset.

\subsection{Phonetic Variant Dictionary (PVD)}
\label{ssec:norm}

Romanized Nepali has no standardized orthography; the same word routinely appears in dozens of orthographic forms \cite{bhat2014normalization,chaudhary2025neptun}. To address this, we construct a \textbf{Phonetic Variant Dictionary (PVD)} -- a lexical resource mapping every observed spelling variant to a single canonical base word.

\paragraph{Rule-based pre-processing.}
Seven deterministic cleaning rules are applied: (i) lowercasing; (ii) Unicode normalization via \texttt{unidecode}; (iii) whitespace normalization; (iv) repeated-letter collapsing ($\geq 3 \to 2$); (v) non-alphanumeric filtering; (vi) word-final \texttt{w}~$\to$~\texttt{a} (non-English only); (vii) \texttt{x}~$\to$~\texttt{chh} (non-English only).

\paragraph{Phonetic clustering.}
We develop a phonetic encoding algorithm adapted from Soundex/Metaphone \cite{philips2000double} and modified for Nepali phonological patterns. Each pre-processed token is normalized (e.g., repeated vowels collapsed, \emph{ae}~$\to$~\emph{ai}), and digraphs such as \emph{chh}/\emph{ch}, \emph{kh}, \emph{bh}, \emph{th}, and \emph{sh} are treated as single units. Consonants are mapped to symbols with aspirated and unaspirated forms merged (the place, voicing, and aspiration classes of Nepali consonants are summarized in Appendix~\ref{sec:phoneme_chart}, Table~\ref{tab:devanagari_wikipedia_format}) (e.g., \emph{k}/\emph{kh}~$\to$~\texttt{K}, \emph{b}/\emph{bh}/\emph{v}~$\to$~\texttt{B}); word-internal vowels are dropped while initial and final vowels are kept, giving a code of at most six symbols (e.g., \emph{huncha}, \emph{hunchha}, \emph{hunachha}~$\to$~\texttt{HNCA}). Tokens sharing the same code are automatically grouped. Clusters were manually validated to correct phonetic collisions (e.g., \emph{bal} and \emph{bol} both~$\to$~\texttt{BL}). The complete encoding rules are included in the released PVD construction code.

\paragraph{Base word assignment.}
Each verified cluster is assigned one \emph{base word}---the most frequent and conventional spelling---serving as the canonical form. Representative PVD entries are shown in Table~\ref{tab:pvd_samples}.

\begin{table}[t]
\centering\small
\setlength{\tabcolsep}{3pt}
\renewcommand{\arraystretch}{1.1}
\begin{tabular}{@{}>{\raggedright\arraybackslash}p{2.1cm}>{\raggedright\arraybackslash}p{5.2cm}@{}}
\toprule
\textbf{Base Word} & \textbf{Variants} \\
\midrule
baachna & bachna, bachnaaa, bachnah, bachanah, bachhanah, baachnah \\
bachaauna & bachauna, bchaauna, bchauna, bhachauna, bachhauna \\
milaaideu & melaideu, milaedeuu, milaideu \\
khuwaayera & khuwayera, khwayera, khuwayra, kwayara, khwwayara \\
\bottomrule
\end{tabular}
\caption{Representative PVD entries showing base words and their spelling variants.}
\label{tab:pvd_samples}
\end{table}

\noindent The final PVD contains \textbf{15,651} canonical base words covering \textbf{92,762} spelling variants.

\subsection{Parallel Corpus Construction and Augmentation}
\label{ssec:parallel}

\paragraph{Manual annotation.}
A total of \textbf{20,500} Romanized sentences were manually transcribed into Devanagari by native Nepali speakers. Each transcription was independently verified for semantic and phonetic accuracy, with 96.2\% accepted without modification; the remaining cases were corrected and adjudicated. Of the input sentences, 80\% came from the cleaned YouTube/Facebook corpus, while 20\% were authored by annotators to cover underrepresented conversational vocabulary.

\paragraph{PVD-based augmentation.}
The 20,500 annotated pairs were split into 19,500 training and 1,000 held-out pairs. Augmentation was applied only to the training set, producing 66,370 filtered variants and 85,870 training pairs in total. The held-out pairs and the separately annotated 500-sentence test set (Section~\ref{ssec:setup}) were kept unchanged. Shared base vocabulary across splits was allowed, as no test-specific variants or annotations were used during training. Table~\ref{tab:corpus} summarizes the final corpus.

\begin{table}[t]
\centering\small
\begin{tabular}{lr}
\toprule
\textbf{Split} & \textbf{Sentence Pairs} \\
\midrule
Original training & 19,500 \\
PVD-augmented training & 66,370 \\
\midrule
\textbf{Total training} & \textbf{85,870} \\
\bottomrule
\end{tabular}
\caption{\system{} training parallel corpus statistics.}
\label{tab:corpus}
\end{table}

\section{System Architecture}
\label{sec:method}

At a high level of our final proposed system, each input sentence passes through
(1)~Rule-Based Normalization, (2)~PVD-based normalization, (3)~\byt{} model fine-tuned on the augmented data, and (4)~a rule-based post-processor (optional). The post-processor restores English tokens to their original Roman form, since they were transcribed into Devanagari during annotation; the procedure is described in Appendix~\ref{ssec:post-process}, and Table~\ref{tab:post-process} traces four code-mixed inputs from raw text to the final output.

\subsection{Inference-Time Normalization}
\label{ssec:pre-process}

Every Romanized input token is first passed through the seven
rule-based cleaning steps described in Section~\ref{ssec:norm}.
The cleaned token is then looked up in the PVD:
\begin{itemize}\setlength\itemsep{2pt}
  \item If the token matches a \emph{base word} entry, it is kept
        unchanged.
  \item If the token matches a \emph{variant} entry, it is replaced
        by the corresponding base word.
  \item If the token is not found in the PVD (e.g., a rare neologism
        or unseen slang), it is passed through unchanged, requiring the model to generalize from byte-level representations.
\end{itemize}
This step substantially reduces the effective vocabulary size seen
by the model and ensures that orthographic variants of the same word
receive equivalent representations during both training and inference.

\subsection{Long-Sequence Chunking}
\label{ssec:chunking}
 
\byt{} operates on UTF-8 byte sequences with a maximum training input length of 300 bytes and output length of 512 bytes. Devanagari outputs are substantially longer in UTF-8 than their Romanized inputs (e.g., ``\emph{bro k xa yar halkhabar}'' is 22 bytes, but its transliteration ``\dv{ब्रो के छ यार हालखबर}'' is 52 bytes), so longer input comments risk exceeding this. To avoid truncation, we apply a lightweight chunking strategy: each normalized sentence is greedily segmented into contiguous chunks of at most 200 characters, preserving word boundaries where possible. Each chunk is transliterated independently and outputs are concatenated in order. Most social-media comments fit within a single chunk; chunking is needed only for unusually long inputs.
 
\subsection{Model Architectures}
\label{ssec:arch}
 
We fine-tune two complementary sequence-to-sequence Transformers \cite{vaswani2017attention} spanning the byte-level versus subword design space, since this axis is central to our research question of which tokenization better handles irregular Romanized Nepali spellings. We use \byt{} rather than other byte/character-level encoders such as CANINE~\cite{clark2022canine}, which is encoder-only and unsuited to sequence generation.
 
\paragraph{ByT5.}
\byt{}~\cite{xue2022byt5} processes raw UTF-8 bytes rather than subword tokens via a full encoder--decoder Transformer, reducing out-of-vocabulary issues for Romanized Nepali's irregular phonetic spellings. Byte-level representations are effective for character-sensitive tasks like transliteration \cite{clark2022canine,yu2023megabyte,xue2022byt5} and are robust to noisy, code-mixed input.
 
\paragraph{mT5.}
\mfive{}~\cite{xue2021mt5} is a multilingual T5~\cite{raffel2020t5} variant pretrained on mC4 (101 languages) with a 250K-vocabulary SentencePiece~\cite{kudo2018sentencepiece} tokenizer covering Nepali, giving a strong language-model prior during decoding. However, its fixed subword tokenizer may fragment unusual Romanized spellings, limiting robustness to highly irregular inputs.

\subsection{Training Configuration}
\label{ssec:training}

We fine-tune \texttt{google/byt5-small} and \texttt{google/mt5-small} on the 85.8k+ PVD-augmented dataset, reserving 1,000 sentence pairs for validation. Both models train on a single NVIDIA A100 GPU for 20 epochs using \texttt{bfloat16} mixed precision. We restrict experiments to the \emph{small} variants of both models because of the computational budget constraints noted in the Limitations; evaluating \emph{base} or larger variants is left to future work.

Training uses AdamW \cite{loshchilov2018decoupled} with a learning rate of $5\!\times\!10^{-5}$, a 500-step warmup, gradient clipping at 1.0, and target pad token masking ($-100$). Inference uses greedy decoding (beam size~1). The only difference is that ByT5 uses byte-level tokenization, while mT5 uses its native SentencePiece tokenizer.
\section{Experiments and Results}
\label{sec:results}

We report quantitative results (Section~\ref{ssec:quant}), comparison against existing tools (Section~\ref{ssec:baselines}), ablation results (Section~\ref{ssec:ablation}), and finally, a detailed error analysis (Section~\ref{ssec:error}) that categorizes recurring failure modes and directly informs the limitations discussed later in the paper.

\subsection{Experimental Setup}
\label{ssec:setup}

\paragraph{Hardware.}
Both \byt{} and \mfive{} are trained
on a single NVIDIA A100 GPU. We evaluate both models under two conditions:
training on the original training data (19,500 sentence pairs) and on the
augmented training data (85,870 sentence pairs; Table~\ref{tab:corpus}).

\paragraph{Evaluation metrics.}
We evaluate all models on a raw social media test set of 500 sentences, annotated separately from the original 20.5k parallel dataset, covering diverse homophones and orthographies. This single test set is used for all results reported in Tables~\ref{tab:quant}--\ref{tab:ablation} and Table~\ref{tab:full_ablations}; the 1,000 held-out pairs of the parallel corpus are used for validation.

We report the following metrics:
\begin{enumerate}\setlength\itemsep{1pt}
  \item \textbf{Character Error Rate (\cer{})} $= \frac{S + D + I}{N}$, where $S$,
        $D$, and $I$ denote substitutions, deletions, and insertions and $N$ is
        the reference length;
  \item \textbf{Word Error Rate (\wer{})}, the word-level equivalent
        \cite{woodard1982speech};
  \item \textbf{Character Accuracy} ($1 - \cer{}$); and
  \item \textbf{Word Accuracy} ($1 - \wer{}$).
\end{enumerate}

\subsection{Quantitative Results}
\label{ssec:quant}

Table~\ref{tab:quant} reports performance across all model variants on the separately annotated 500-sentence test set (Section~\ref{ssec:setup}).

\begin{table}[t]
\centering\small
\setlength{\tabcolsep}{3.5pt}
\renewcommand{\arraystretch}{1.1}
\begin{tabular}{llcccc}
\toprule
\textbf{Model} & \textbf{Data} & \textbf{CER} & \textbf{WER} & \textbf{\makecell{Char.\\Acc.}} & \textbf{\makecell{Word\\Acc.}} \\
\midrule
\multirow{2}{*}{\byt{}}
  & Aug.  & \textbf{5.96} & \textbf{21.85} & \textbf{94.04} & \textbf{78.15} \\
  & Orig. & 6.03 & 23.23 & 93.97 & 76.77 \\
\midrule
\multirow{2}{*}{\mfive{}}
  & Aug.  & 13.50 & 31.52 & 86.50 & 68.48 \\
  & Orig. & 17.45 & 37.20 & 82.55 & 62.80 \\
\bottomrule
\end{tabular}
\caption{Performance on the separately annotated 500-sentence social-media test set (the same test set as in Table~\ref{tab:baselines}).
Aug.\ = dictionary-augmented training; Orig.\ = original training.
All error rates and accuracies are reported in percentages.}
\label{tab:quant}
\end{table}

Data augmentation consistently reduced error rates across both architectures, with \byt{} on augmented data performing best (5.96\% CER, 21.85\% WER). The gain differs by architecture: \mfive{} improves by 3.95 CER and 5.68 WER points, whereas \byt{} improves by only 0.07 CER (6.03\%$\to$5.96\%) and 1.38 WER points (23.23\%$\to$21.85\%). The \byt{} CER difference is small enough to fall within run-to-run variation; we report no variance or significance tests. Appendix~\ref{sec:ablations} (Table~\ref{tab:full_ablations}) suggests that inference-time normalization and augmentation draw on the same PVD and thus address overlapping spelling variation, leaving little headroom for \byt{}. Example outputs from all four model variants are shown in Appendix~\ref{sec:model_comparison} (Table~\ref{tab:model_comparison}).

Two factors may explain these results:
\begin{enumerate}
    \item \textbf{Byte-level superiority:} \byt{} outperforms \mfive{} in all settings on our test set, suggesting that byte-level modeling better handles the irregular orthography of Romanized Nepali.
    \item \textbf{Subword sensitivity:} \mfive{} struggles with non-standard spellings, plausibly due to its SentencePiece subword segmentation.
\end{enumerate}

\subsection{Comparison with Existing Tools}
\label{ssec:baselines}

Table~\ref{tab:baselines} compares our best model
(\byt{} trained on augmented dataset) against publicly available
Nepali transliteration systems on the same 500-sentence social-media test set used in Table~\ref{tab:quant} (Section~\ref{ssec:setup}).
These 500 sentences were newly sampled from social media comments and annotated separately from the 20.5k parallel corpus.

\begin{table}[t]
\centering
\small
\setlength{\tabcolsep}{3pt}
\renewcommand{\arraystretch}{1.10}
\begin{tabular}{@{}lcc@{}}
\toprule
\textbf{System} & \textbf{\cer{}} & \textbf{\wer{}} \\
\midrule
\byt{} Norm. (Ours)     & \textbf{5.96} & \textbf{21.85} \\
\byt{} Raw (Ours)       & 7.24          & 24.39 \\
NepaliXlit \cite{patel2026from}
                        & 9.22         & 43.68 \\
Google Transliteration \cite{googletranslit}
                        & 20.76         & 60.88 \\
Easy Nepali Typing \citeyearpar{easynepalityping}
                        & 21.22         & 61.56 \\
Ashesh Unicode \cite{asheshunicode}
                        & 21.93         & 69.07 \\
Kokil Unicode \cite{kokilunicode}
                        & 27.51         & 78.45 \\
Nepali Unicode \citeyearpar{nepaliunicodecom}
                        & 30.87         & 82.71 \\
\bottomrule
\end{tabular}
\caption{Comparison with existing transliteration systems on the 500-sentence
social-media test set.
Norm.\ = model output with normalization;
Raw = without pre-processing.
All error rates are reported in percentages.}
\label{tab:baselines}
\end{table}

The normalized \byt{} model achieves the lowest CER (5.96\%) and WER (21.85\%), substantially outperforming NepaliXlit, particularly in WER. Removing normalization increases CER and WER to 7.24\% and 24.39\%, respectively, showing the benefit of the proposed preprocessing. Notably, even the raw \byt{} model outperforms all external systems, indicating that the gains are not due to preprocessing alone. Table~\ref{tab:existing_systems_qualitative} (Appendix~\ref{sec:existing_systems_qualitative}) gives qualitative comparisons on two inputs: the rule-based tools often mis-render English loanwords (e.g., \emph{vlog} as \dv{व्लोग}) and drop vowel diacritics, whereas our normalized model closely matches the reference.

\subsection{Ablation Results}
\label{ssec:ablation}
We ablate the pre-processing steps on the \byt{} augmented model (Table~\ref{tab:ablation}). Either normalization alone improves over raw input (CER 6.46\% for PVD, 6.65\% for rule-based, vs.\ 7.24\% raw), and combining both gives the best results (CER 5.96\%, WER 21.85\%), so the two are complementary. The same ordering holds for \byt{} Original and \mfive{} Original, while for \mfive{} Augmented rule-based alone is slightly better than PVD alone (14.04\% vs.\ 14.28\% CER). For all four models, removing both steps causes the largest degradation; full results are in Table~\ref{tab:full_ablations} (Appendix~\ref{sec:ablations}).

\begin{table}[t]
\centering\small
\setlength{\tabcolsep}{4pt}
\begin{tabular}{@{}lcccc@{}}
\toprule
\textbf{Configuration} & \textbf{CER} & \textbf{WER} & \textbf{\makecell{Char.\\Acc.}} & \textbf{\makecell{Word\\Acc.}} \\
\midrule
\textbf{Rule-Based + PVD} & \textbf{5.96} & \textbf{21.85} & \textbf{94.04} & \textbf{78.15} \\
PVD only                  & 6.46 & 22.32 & 93.54 & 77.68 \\
Rule-Based only           & 6.65 & 23.55 & 93.35 & 76.45 \\
Raw (no normalization)    & 7.24 & 24.39 & 92.76 & 75.61 \\
\bottomrule
\end{tabular}
\caption{Ablation results for the \byt{} augmented model across pre-processing configurations.}
\label{tab:ablation}
\end{table}

\subsection{Human Evaluation}
\label{ssec:human}

To complement quantitative evaluation, the model output of our proposed model \textbf{ByT5-Augmented} for 500 raw Romanized Nepali sentences taken from social media comments; as in Section~\ref{ssec:baselines}, were evaluated by ten native Nepali speakers. Each transliterated output produced by the proposed
\byt{} Augmented model was rated on a 5-point Likert scale using two
criteria:

\begin{itemize}
    \item \textbf{Accuracy}: Whether the Romanized input is correctly
    transliterated into the intended Devanagari words.
    \item \textbf{Naturalness}: Whether the transliterated sentence
    reads like naturally written Nepali.
\end{itemize}
For example, given the Romanized input
\emph{``ma ghar jadai xu''}, meaning \emph{``I am going home''}, the output
``\dv{म घर जादै छु}'' would be considered accurate but less natural,
whereas ``\dv{म घर जाँदै छु}'' (with nasalization marker `\dv{ँ}') would be both accurate and natural.

Table~\ref{tab:human} presents the average ratings.

\begin{table}[t]
\centering\small
\renewcommand{\arraystretch}{1.15}
\begin{tabular}{lcc}
\toprule
\textbf{Criterion} & \textbf{Mean ($\pm$Std)} & \textbf{Scale} \\
\midrule
Accuracy     & 4.61 $\pm$ 0.61 & 1--5 \\
Naturalness  & 4.35 $\pm$ 0.67 & 1--5 \\
\midrule
Average      & 4.48 &   \\
\bottomrule
\end{tabular}
\caption{Human evaluation results for \byt{} Augmented (500~samples,
10~raters). Scale: 1~=~very poor, 5~=~excellent. The detailed annotation guidelines are provided in Appendix~\ref{sec:likert_guidelines}.}
\label{tab:human}
\end{table} 

The proposed model achieved an average Accuracy score of 4.61 and a Naturalness score of 4.35,
resulting in an overall mean rating of 4.48 out of 5. The high scores indicate that the generated
transliterations are generally correct while remaining natural and readable for native Nepali
speakers. Accuracy received slightly higher ratings than Naturalness, suggesting that although the
intended words are usually recovered correctly, minor variations in spelling or word choice
occasionally reduce perceived naturalness.

A two-tailed paired t-test indicates a small but statistically
significant difference between Accuracy and Naturalness ($p < 0.01$),
suggesting that outputs are slightly more accurate than fluent.
The narrow standard deviations ($\leq 0.67$) indicate consistent ratings
across annotators, and severe errors are rare.

\subsection{Error Analysis}
\label{ssec:error}

Inspection of low-scoring outputs reveals three recurring error categories.

\paragraph{(1) Residual orthographic variation.}
Despite augmentation, the model occasionally produces different
Devanagari outputs for phonetically equivalent input spellings.
For example, \emph{halkhabar}~$\to$~\dv{हालखबर} (correct) but
\emph{haalkhabar}~$\to$~\dv{हलखबर} (incorrect).
This suggests the model has not fully generalized vowel
distinctions across all word contexts.

\paragraph{(2) Homophonic ambiguity.}
Romanized strings mapping to multiple Devanagari words cause semantic
errors:
\emph{kala}~$\to$~\dv{कला} (art) vs.\ \dv{काला} (black);
\emph{bal}~$\to$~\dv{बल} (strength) vs.\ \dv{बाल} (hair);
\emph{tato}~$\to$~\dv{तातो} (hot) vs.\ \dv{टाटो} (mark).
Resolving these requires discourse context beyond the sentence level, a
known challenge in low-resource transliteration \cite{roark2020dakshina}.

\paragraph{(3) Diacritic errors.}
Four sub-types are observed:
(a)~\emph{chandrabindu} (\dv{ँ}) vs.\ \emph{anusvara} (\dv{◌ं})
confusion e.g., \emph{sanga}~$\to$~\dv{संग} instead of \dv{सँग} (these symbols relate to the nasalization of words);
(b)~short/long vowel confusion (\dv{इ}/\dv{ई}, \dv{उ}/\dv{ऊ})
e.g., \emph{ful}~$\to$~\dv{फुल} instead of \dv{फूल};
(c)~\emph{visarga} (\dv{ः}) omission
e.g., \emph{namah}~$\to$~\dv{नम} instead of \dv{नमः};
(d)~\emph{halant} (\dv{्}) omission
e.g., \emph{hun}~$\to$~\dv{हुन} instead of \dv{हुन्‌} (this symbol suppresses the vowel sound of a consonant to create a half-letter sound.)

Overall, the remaining errors primarily arise from fine-grained orthographic distinctions and context-dependent mappings. These findings highlight areas for further improvement in data coverage and context-aware modeling.
\section{Conclusion}
\label{sec:conclusion}

We presented \textbf{\system{}}, a low-resource system for transliterating code-mixed Romanized Nepali-English text to Devanagari script, combining large-scale data collection, a Phonetic Variant Dictionary for normalization and augmentation, a manually annotated parallel corpus (20,500 pairs, expanded to 85,870), and fine-tuned \byt{} and \mfive{} models. Our best model (\byt{}, augmented) achieves \cer{}~=~5.96\% and \wer{}~=~21.85\%, outperforming existing Nepali transliteration tools.

Future work includes expanding the PVD to cover more dialects and evolving slangs, scaling the parallel corpus via semi-supervised methods (e.g., back-translation, self-training), exploring paragraph-level decoding for discourse-level ambiguities, and training a multilingual Devanagari-English language model for downstream tasks.

Beyond Nepali, our pipeline is largely language-agnostic---the PVD's phonetic-clustering algorithm and the augmentation/chunking strategies make no 
Nepali-specific assumptions and should transfer readily
to other Romanized low-resource languages (e.g., Sinhala-English, 
Bengali-English) with only the phonetic encoding rules needing re-tuning.

\section*{Limitations}
\label{sec:limitations}

\paragraph{Static normalization coverage.}
The PVD is a fixed resource and cannot generalize to unseen spelling
variants, neologisms, or emerging slang.

\paragraph{Diacritic and homophonic errors.}
As analyzed in Section~\ref{ssec:error}, the model systematically
confuses \emph{anusvara}~(\dv{◌ं}) vs.~\emph{chandrabindu}~(\dv{◌ँ}),
short/long vowels (\dv{इ}/\dv{ई}, \dv{उ}/\dv{ऊ}), and omits
\emph{halant}~(\dv{्}) and \emph{visarga}~(\dv{ः}) in some contexts.
Homophonic ambiguities require discourse-level context that our
sentence-level model lacks---a fundamental architectural constraint.

\paragraph{English loanword transliteration.}
The model performs poorly on English loanwords (e.g., \emph{static},
\emph{quantify}) due to training data constraints and the absence of
code-switching awareness, yielding unnatural transliterations.

\paragraph{Low-resource constraints.}
The modest size and diversity of our training corpus limit
exposure to the full range of Nepali romanization conventions and
writing styles. Computational
budget limits hyperparameter optimization, ensemble
methods, or training large-size models. We release our resources
publicly to facilitate future work.

\section*{Ethical Considerations}
\label{sec:ethical}
Our dataset is built from publicly available YouTube and Facebook comments. Human annotators were informed, consented, and compensated at or above local minimum wage. The system is intended for research in Nepali NLP and not for surveillance or harmful profiling. Outputs may contain errors and should be verified in high-stakes use cases. No private information is released.

\section*{Social Impact}
\label{sec:social}
Reliable transliteration of code-mixed Romanized Nepali-English text can improve access to Nepali NLP tools such as sentiment analysis, hate-speech detection, and information retrieval, supporting content moderation, accessibility, and language-technology inclusion for an under-resourced language community. However, transliteration errors may propagate to downstream systems and cause incorrect predictions or moderation decisions. We therefore recommend human review for moderation and other safety-critical applications.

\section*{Acknowledgments}
This work was supported by the University Grants Commission, Nepal
under Collaborative Research Grant (UGC Award No.\ CRG-81/82-Engg-01)
for the project ``Analysis of Sentiments and Insights from Nepali
Social Media and News Contents for Data-Driven Governance.''
We thank our annotators and human evaluators for their careful work.

\bibliography{lipimix}

@misc{eberhard2023ethnologue,
  author    = {Eberhard, David M. and Simons, Gary F. and Fennig, Charles D.},
  title     = {Ethnologue: Languages of the World},
  edition   = {26th},
  year      = {2023},
  publisher = {SIL International},
  address   = {Dallas, Texas},
  url       = {http://www.ethnologue.com}
}

@article{chaudhary2025neptun,
  author    = {Chaudhary, Chandra Prakash and Joshi, Basanta and
               Shakya, Aman and Giri, Santosh},
  title     = {{NEPTUN}: Normalization for {R}omanized {N}epali Sentiment
               Analysis},
  journal   = {Journal of Himalaya College of Engineering},
  volume    = {2},
  number    = {1},
  pages     = {19--28},
  year      = {2025},
  publisher = {Nepal Journals Online (JOL)},
  doi       = {10.3126/jhcoe.v2i1.91508}
}

@inproceedings{patel2026from,
  author    = {Patel, Suraj and Dhami, Kashish Kumari and
               Sherpa, Norden and Khadka, Supriya},
  title     = {From {R}omanized to {D}evanagari: Enhancing {N}epali Sentiment
               Analysis with {N}epali{X}lit},
  booktitle = {Proceedings of the Second Workshop on Challenges in Processing
               South Asian Languages (CHiPSAL 2026)},
  year      = {2026},
  month     = may,
  publisher = {European Language Resources Association},
  doi       = {10.13140/RG.2.2.23900.35202}
}

@misc{kafley2024transliteration,
  author       = {Kafley, Saugat},
  title        = {{N}epali--{R}oman Transliteration Dataset},
  year         = {2024},
  howpublished = {Hugging Face Datasets},
  note         = {Word-level {R}omanized--{D}evanagari pairs.
                  Available at
                  \url{https://huggingface.co/datasets/Saugatkafley/Nepali-Roman-Transliteration}.
                  Accessed: 2026-05-10}
}

@misc{pandey2025codemixed,
  author       = {Pandey, Manish and Yadav, Nageshwar Prasad and
                  Adduru, Mokshada and Rai, Sawan},
  title        = {Creating and Evaluating Code-Mixed {N}epali--{E}nglish and
                  {T}elugu--{E}nglish Datasets for Abusive Language Detection
                  Using Traditional and Deep Learning Models},
  year         = {2025},
  eprint       = {2504.21026},
  archivePrefix = {arXiv},
  primaryClass = {cs.CL},
  url          = {https://arxiv.org/abs/2504.21026}
}

@misc{bewakes_unicode,
  author = {Bewakes},
  title  = {Nepali {Unicode} Romanized to {Devanagari} Converter},
  year   = {2021},
  url    = {https://github.com/bewakes/nepali-unicode-converter}
}

@misc{asheshunicode,
  author = {Ashesh},
  title = {Nepali Unicode: Romanized to Unicode Converter},
  howpublished = {Online},
  note = {Available at \url{https://www.ashesh.com.np/nepali-unicode.php}. Accessed: 2026-05-10},
  year = {2026}
}

@misc{kokilunicode,
  author = {Kokil},
  title = {Romanized Nepali to Unicode Converter},
  howpublished = {Online},
  note = {Available at \url{https://kokil.com.np/tools/romanized-nepali-to-unicode}. Accessed: 2026-05-10},
  year = {2026}
}

@misc{googletranslit,
  author = {Google},
  title = {Google Transliteration},
  howpublished = {Online},
  note = {Available at \url{https://www.google.com/intl/ne/inputtools/try/}. Accessed: 2026-05-10},
  year = {2026}
}

@misc{easynepalityping,
  author = {{Easy Nepali Typing}},
  title = {Easy Nepali Typing},
  howpublished = {Online},
  note = {Available at \url{https://www.easynepalityping.com/}. Accessed: 2026-05-10},
  year = {2026}
}

@misc{nepaliunicodecom,
  author = {{Nepali Unicode}},
  title = {Nepali Unicode Typing},
  howpublished = {Online},
  note = {Available at \url{https://www.nepali-unicode.com}. Accessed: 2026-05-10},
  year = {2026}
}

@misc{wikipedia_devanagari,
    author       = {Wikipedia},
    title        = {Devanagari},
    howpublished = {Online},
    note         = {Available at \url{https://en.wikipedia.org/wiki/Devanagari}. Accessed: 2026-05-10},
    year         = {2026},
}

@inproceedings{dogruoz2021survey,
  author    = {Do\u{g}ru\"{o}z, A.~Seza and Sitaram, Sunayana and
               Bullock, Barbara E. and Toribio, Almeida~Jacqueline},
  title     = {A Survey of Code-Switching: Linguistic and Social Perspectives
               for Language Technologies},
  booktitle = {Proceedings of the 59th Annual Meeting of the Association for
               Computational Linguistics and the 11th International Joint
               Conference on Natural Language Processing (Volume 1: Long Papers)},
  year      = {2021},
  pages     = {1654--1666},
  publisher = {Association for Computational Linguistics},
  doi       = {10.18653/v1/2021.acl-long.131}
}

@inproceedings{winata2023codeswitching,
  author    = {Winata, Genta~Indra and Aji, Alham~Fikri and
               Yong, Zheng~Xin and Solorio, Thamar},
  title     = {The Decades Progress on Code-Switching Research in {NLP}:
               A Systematic Survey on Trends and Challenges},
  booktitle = {Findings of the Association for Computational Linguistics:
               ACL 2023},
  year      = {2023},
  pages     = {2936--2978},
  publisher = {Association for Computational Linguistics},
  doi       = {10.18653/v1/2023.findings-acl.185}
}

@inproceedings{das2014identifying,
  author    = {Das, Amitava and Gamb\"{a}ck, Bj\"{o}rn},
  title     = {Identifying Languages at the Word Level in Code-Mixed {I}ndian
               Social Media Text},
  booktitle = {Proceedings of the 11th International Conference on Natural
               Language Processing},
  year      = {2014},
  pages     = {378--387},
  address   = {Goa, India},
  publisher = {NLP Association of India}
}

@inproceedings{srivastava2020phinc,
  author    = {Srivastava, Vivek and Singh, Mayank},
  title     = {{PHINC}: A Parallel {H}inglish Social Media Code-Mixed Corpus
               for Machine Translation},
  booktitle = {Proceedings of the Sixth Workshop on Noisy User-Generated Text
               (W-NUT 2020)},
  year      = {2020},
  pages     = {41--49},
  publisher = {Association for Computational Linguistics},
  doi       = {10.18653/v1/2020.wnut-1.7}
}

@inproceedings{pratapa2018language,
  author    = {Pratapa, Adithya and Bhat, Gayatri and Choudhury, Monojit and
               Sitaram, Sunayana and Dandapat, Sandipan and Bali, Kalika},
  title     = {Language Modeling for Code-Mixing: The Role of Linguistic Theory
               Based Synthetic Data},
  booktitle = {Proceedings of the 56th Annual Meeting of the Association for
               Computational Linguistics (Volume 1: Long Papers)},
  year      = {2018},
  pages     = {1543--1553},
  publisher = {Association for Computational Linguistics},
  doi       = {10.18653/v1/P18-1143}
}

@inproceedings{mave2018language,
  author    = {Mave, Deepthi and Maharjan, Suraj and Solorio, Thamar},
  title     = {Language Identification and Analysis of Code-Switched Social
               Media Text},
  booktitle = {Proceedings of the Third Workshop on Computational Approaches
               to Linguistic Code-Switching},
  year      = {2018},
  pages     = {51--61},
  publisher = {Association for Computational Linguistics},
  doi       = {10.18653/v1/W18-3206}
}

@inproceedings{rijhwani2017estimating,
  author    = {Rijhwani, Shruti and Sequiera, Royal and Choudhury, Monojit and
               Bali, Kalika and Maddila, Chandra~Shekhar},
  title     = {Estimating Code-Switching on {T}witter with a Novel Generalized
               Word-Level Language Detection Technique},
  booktitle = {Proceedings of the 55th Annual Meeting of the Association for
               Computational Linguistics (Volume 1: Long Papers)},
  year      = {2017},
  pages     = {1971--1982},
  publisher = {Association for Computational Linguistics},
  doi       = {10.18653/v1/P17-1180}
}

@inproceedings{kartik2024synthetic,
  author    = {Kartik and Soni, Sanjana and Kunchukuttan, Anoop and
               Chakraborty, Tanmoy and Akhtar, Md~Shad},
  title     = {Synthetic Data Generation and Joint Learning for Robust
               Code-Mixed Translation},
  booktitle = {Proceedings of the 2024 Joint International Conference on
               Computational Linguistics, Language Resources and Evaluation
               (LREC-COLING 2024)},
  year      = {2024},
  pages     = {15480--15492},
  publisher = {ELRA and ICCL}
}

@inproceedings{hamed2023augmentation,
  author    = {Hamed, Injy and Habash, Nizar and Vu, Ngoc Thang},
  title     = {Data Augmentation Techniques for Machine Translation of
               Code-Switched Texts: A Comparative Study},
  booktitle = {Findings of the Association for Computational Linguistics:
               EMNLP 2023},
  year      = {2023},
  pages     = {140--154},
  publisher = {Association for Computational Linguistics},
  doi       = {10.18653/v1/2023.findings-emnlp.11}
}

@inproceedings{wang2025codeswitching,
  author    = {Wang, Zhijun and Li, Jiahuan and Zhou, Hao and Weng, Rongxiang
               and Wang, Jingang and Huang, Xin and Han, Xue and Feng, Junlan
               and Deng, Chao and Huang, Shujian},
  title     = {Investigating and Scaling up Code-Switching for Multilingual
               Language Model Pre-Training},
  booktitle = {Findings of the Association for Computational Linguistics:
               ACL 2025},
  year      = {2025},
  pages     = {11032--11046},
  address   = {Vienna, Austria},
  publisher = {Association for Computational Linguistics}
}

@inproceedings{madhani2023aksharantar,
  author    = {Madhani, Yash and Parthan, Sushane and Bedekar, Priyanka and
               Nc, Gokul and Khapra, Ruchi and Kunchukuttan, Anoop and
               Kumar, Pratyush and Khapra, Mitesh},
  title     = {Aksharantar: Open {I}ndic-language Transliteration Datasets
               and Models for the Next Billion Users},
  booktitle = {Findings of the Association for Computational Linguistics:
               EMNLP 2023},
  year      = {2023},
  pages     = {40--57},
  publisher = {Association for Computational Linguistics},
  doi       = {10.18653/v1/2023.findings-emnlp.4}
}

@inproceedings{kunchukuttan2021largescale,
  author    = {Kunchukuttan, Anoop and Jain, Siddharth and Kejriwal, Rahul},
  title     = {A Large-Scale Evaluation of Neural Machine Transliteration for
               {I}ndic Languages},
  booktitle = {Proceedings of the 16th Conference of the European Chapter of
               the Association for Computational Linguistics: Main Volume},
  year      = {2021},
  pages     = {3469--3475},
  publisher = {Association for Computational Linguistics},
  doi       = {10.18653/v1/2021.eacl-main.303}
}

@inproceedings{roark2020dakshina,
  author    = {Roark, Brian and Wolf-Sonkin, Lawrence and
               Kirov, Christo and Mielke, Sabrina~J. and Johny, Cibu and
               Demir\c{s}ahin, I\c{s}{\i}n and Hall, Keith},
  title     = {Processing {S}outh {A}sian Languages Written in the {L}atin
               Script: the {D}akshina Dataset},
  booktitle = {Proceedings of the 12th Language Resources and Evaluation Conference},
  year      = {2020},
  pages     = {2413--2423},
  publisher = {European Language Resources Association}
}

@inproceedings{bhat2014normalization,
  author    = {Bhat, Irshad~Ahmad and Mujadia, Vandan and Tammewar, Aniruddha
               and Bhat, Riyaz~Ahmad and Shrivastava, Manish},
  title     = {{IIIT-H} System Submission for {FIRE}2014 Shared Task on
               Transliterated Search},
  booktitle = {Proceedings of the Forum for Information Retrieval Evaluation
               (FIRE 2014)},
  year      = {2014},
  pages     = {48--53},
  publisher = {ACM},
  doi       = {10.1145/2824864.2824872}
}

@inproceedings{gella2013query,
  author    = {Gella, Spandana and Sharma, Jatin and Bali, Kalika},
  title     = {Query Word Labeling and Back Transliteration for {I}ndian
               Languages: Shared Task System Description},
  booktitle = {FIRE 2013 Working Notes, Forum for Information Retrieval
               Evaluation},
  year      = {2013}
}

@inproceedings{amrhein2020romanization,
  author    = {Amrhein, Chantal and Sennrich, Rico},
  title     = {On Romanization for Model Transfer Between Scripts in Neural
               Machine Translation},
  booktitle = {Findings of the Association for Computational Linguistics:
               EMNLP 2020},
  year      = {2020},
  pages     = {2461--2469},
  publisher = {Association for Computational Linguistics},
  doi       = {10.18653/v1/2020.findings-emnlp.223}
}

@article{xue2022byt5,
  author    = {Xue, Linting and Barua, Aditya and Constant, Noah and
               Al-Rfou, Rami and Narang, Sharan and Kale, Mihir and
               Roberts, Adam and Raffel, Colin},
  title     = {{ByT5}: Towards a Token-Free Future with Pre-Trained
               Byte-to-Byte Models},
  journal   = {Transactions of the Association for Computational Linguistics},
  volume    = {10},
  pages     = {291--306},
  year      = {2022},
  doi       = {10.1162/tacl_a_00461}
}

@inproceedings{xue2021mt5,
  author    = {Xue, Linting and Constant, Noah and Roberts, Adam and
               Kale, Mihir and Al-Rfou, Rami and Siddhant, Aditya and
               Barua, Aditya and Raffel, Colin},
  title     = {{mT5}: A Massively Multilingual Pre-Trained Text-to-Text
               Transformer},
  booktitle = {Proceedings of the 2021 Conference of the North American Chapter
               of the Association for Computational Linguistics: Human Language
               Technologies},
  year      = {2021},
  pages     = {483--498},
  publisher = {Association for Computational Linguistics},
  doi       = {10.18653/v1/2021.naacl-main.41}
}

@article{raffel2020t5,
  author    = {Raffel, Colin and Shazeer, Noam and Roberts, Adam and
               Lee, Katherine and Narang, Sharan and Matena, Michael and
               Zhou, Yanqi and Li, Wei and Liu, Peter~J.},
  title     = {Exploring the Limits of Transfer Learning with a Unified
               Text-to-Text Transformer},
  journal   = {Journal of Machine Learning Research},
  volume    = {21},
  number    = {140},
  pages     = {1--67},
  year      = {2020}
}

@inproceedings{vaswani2017attention,
  author    = {Vaswani, Ashish and Shazeer, Noam and Parmar, Niki and
               Uszkoreit, Jakob and Jones, Llion and Gomez, Aidan~N. and
               Kaiser, {\L}ukasz and Polosukhin, Illia},
  title     = {Attention Is All You Need},
  booktitle = {Advances in Neural Information Processing Systems},
  volume    = {30},
  year      = {2017},
  publisher = {Curran Associates, Inc.}
}

@inproceedings{kudo2018sentencepiece,
  author    = {Kudo, Taku and Richardson, John},
  title     = {{SentencePiece}: A Simple and Language Independent Subword
               Tokenizer and Detokenizer for Neural Text Processing},
  booktitle = {Proceedings of the 2018 Conference on Empirical Methods in
               Natural Language Processing: System Demonstrations},
  year      = {2018},
  pages     = {66--71},
  publisher = {Association for Computational Linguistics},
  doi       = {10.18653/v1/D18-2012}
}

@article{clark2022canine,
  author    = {Clark, Jonathan~H. and Garrette, Dan and Turc, Iulia and
               Wieting, John},
  title     = {{CANINE}: Pre-Training an Efficient Tokenization-Free Encoder
               for Language Representation},
  journal   = {Transactions of the Association for Computational Linguistics},
  volume    = {10},
  pages     = {73--91},
  year      = {2022},
  doi       = {10.1162/tacl_a_00448}
}

@inproceedings{yu2023megabyte,
  author    = {Yu, Lili and Simig, D\'{a}niel and Flaherty, Colin and
               Aghajanyan, Armen and Zettlemoyer, Luke and Lewis, Mike},
  title     = {{MEGABYTE}: Predicting Million-Byte Sequences with Multiscale
               Transformers},
  booktitle = {Advances in Neural Information Processing Systems},
  volume    = {36},
  year      = {2023},
  publisher = {Curran Associates, Inc.}
}

@inproceedings{conneau2020unsupervised,
  author    = {Conneau, Alexis and Khandelwal, Kartikay and Goyal, Naman and
               Chaudhary, Vishrav and Wenzek, Guillaume and Guzm\'{a}n, Francisco
               and Grave, \'{E}douard and Ott, Myle and Zettlemoyer, Luke and
               Stoyanov, Veselin},
  title     = {Unsupervised Cross-Lingual Representation Learning at Scale},
  booktitle = {Proceedings of the 58th Annual Meeting of the Association for
               Computational Linguistics},
  year      = {2020},
  pages     = {8440--8451},
  publisher = {Association for Computational Linguistics},
  doi       = {10.18653/v1/2020.acl-main.747}
}

@inproceedings{loshchilov2018decoupled,
  author    = {Loshchilov, Ilya and Hutter, Frank},
  title     = {Decoupled Weight Decay Regularization},
  booktitle = {International Conference on Learning Representations},
  year      = {2019}
}

@article{philips2000double,
  author    = {Philips, Lawrence},
  title     = {The Double Metaphone Search Algorithm},
  journal   = {C/C++ Users Journal},
  volume    = {18},
  number    = {6},
  pages     = {38--43},
  year      = {2000}
}

@inproceedings{woodard1982speech,
  author    = {Woodard, J.~P. and Nelson, J.~T.},
  title     = {An Information Theoretic Measure of Speech Recognition
               Performance},
  booktitle = {Workshop on Standardisation for Speech {I/O} Technology},
  address   = {Naval Air Development Center, Warminster, PA},
  year      = {1982}
}

\clearpage
\onecolumn
\raggedbottom
\appendix

\section{Dataset Collection Statistics}
\label{sec:data_collection}

Table~\ref{tab:raw_comments} summarizes the raw comments collected from each YouTube channel used in this study.

\begin{table}[H]
\centering
\small
\setlength{\tabcolsep}{10pt}
\renewcommand{\arraystretch}{1.15}
\begin{tabular}{@{}lr@{}}
\toprule
\textbf{Channel} & \textbf{Comments} \\
\midrule
Highlights Nepal     & 746,449 \\
MusicNepalPvtLtd     & 657,972 \\
OSR Digital          & 438,271 \\
KantipurTV           & 245,871 \\
TheCartoonCrewz      & 197,543 \\
Indepth Story        & 123,035 \\
ProjectKura          & 107,823 \\
TheNepaliComment     & 103,754 \\
Ideapreneur Nepal    & 97,234  \\
TechlekhYT           & 54,567  \\
\midrule
\textbf{Total}       & \textbf{2,772,519} \\
\bottomrule
\end{tabular}
\caption{Raw comments collected from each YouTube channel, sorted by volume.}
\label{tab:raw_comments}
\end{table}

\section{Nepali Consonant Phoneme Chart}
\label{sec:phoneme_chart}

Table~\ref{tab:devanagari_wikipedia_format} presents the classification of Nepali consonant phonemes by place and manner of articulation, used as reference for the phonetic transliteration scheme.

\begin{table}[H]
\centering
\footnotesize
\setlength{\tabcolsep}{4pt}
\renewcommand{\arraystretch}{1.3}
\begin{tabular}{@{}l ccccccc@{}}
\toprule
\textbf{Phonetics $\rightarrow$} & \multicolumn{4}{c}{\textbf{sparśa (Occlusive / Plosive)}} & \textbf{anunāsika} & \textbf{antastha} & \textbf{ūṣman} \\
\cmidrule(lr){2-5}
\textbf{Voicing $\rightarrow$} & \multicolumn{2}{c}{\textbf{aghoṣa (Voiceless)}} & \multicolumn{2}{c}{\textbf{saghoṣa (Voiced)}} & \textbf{(Nasal)} & \textbf{(Approximant)} & \textbf{(Fricative)} \\
\cmidrule(lr){2-3}\cmidrule(lr){4-5}
\textbf{Aspiration $\rightarrow$} & \textbf{alpaprāṇa} & \textbf{mahāprāṇa} & \textbf{alpaprāṇa} & \textbf{mahāprāṇa} & \textbf{saghoṣa} & \textbf{saghoṣa} & \textbf{aghoṣa} \\
\midrule
\textbf{kaṇṭhya} (Velar) & k (\dv{क}) & k\textsuperscript{h} (\dv{ख}) & g (\dv{ग}) & g\textsuperscript{h} (\dv{घ}) & ŋ (\dv{ङ}) & -- & h (\dv{ह})\textsuperscript{*} \\
\textbf{tālavya} (Palatal) & c (\dv{च}) & c\textsuperscript{h} (\dv{छ}) & j (\dv{ज}) & j\textsuperscript{h} (\dv{झ}) & \~{n} (\dv{ञ}) & y (\dv{य}) & \'{s} (\dv{श}) \\
\textbf{mūrdhanya} (Retroflex) & \d{t} (\dv{ट}) & \d{t}\textsuperscript{h} (\dv{ठ}) & \d{d} (\dv{ड}) & \d{d}\textsuperscript{h} (\dv{ढ}) & \d{n} (\dv{ण}) & r (\dv{र}) & \d{s} (\dv{ष}) \\
\textbf{dantya} (Dental) & t (\dv{त}) & t\textsuperscript{h} (\dv{थ}) & d (\dv{द}) & d\textsuperscript{h} (\dv{ध}) & n (\dv{न}) & l (\dv{ल}) & s (\dv{स}) \\
\textbf{oṣṭhya} (Labial) & p (\dv{प}) & p\textsuperscript{h} (\dv{फ}) & b (\dv{ब}) & b\textsuperscript{h} (\dv{भ}) & m (\dv{म}) & v (\dv{व}) & -- \\
\bottomrule
\end{tabular}
\caption{Devanagari consonant classification mapping linguistic placement, voicing, and aspiration according to traditional phonetics \cite{wikipedia_devanagari}.}
\label{tab:devanagari_wikipedia_format}
\end{table}

\section{Post-processing: English Token Restoration}
\label{ssec:post-process}

Since English words in natural Nepali social media are typically
preserved in Roman script, we apply a rule-based post-processing step
to restore them in the final output.

During annotation, English tokens were converted into Devanagari
phonetic form (e.g., \emph{school}~$\to$~\dv{स्कुल}) to maintain
annotation consistency.

To recover original usage, the post-processor:
(i) detects English tokens in the original input,
(ii) aligns them with corresponding spans in the model output using
dictionary-based and heuristic matching, and
(iii) restores them to their original Roman surface form.

English numerals are also converted to Devanagari digits
(\dv{०}--\dv{९}) for script consistency.

This step introduces no additional training and runs in linear time
$O(n)$. Table~\ref{tab:post-process} traces four code-mixed inputs
through this procedure.

\begin{table}[H]
\centering
\small
\setlength{\tabcolsep}{4pt}
\renewcommand{\arraystretch}{1.3}
\begin{tabular}{c >{\raggedright\arraybackslash}p{3.2cm} >{\raggedright\arraybackslash}p{3.2cm} >{\raggedright\arraybackslash}p{3.8cm} >{\raggedright\arraybackslash}p{3.8cm}}
\toprule
\textbf{\#} & \textbf{Input (Romanized)} & \textbf{Pre-processed} & \textbf{Raw Model Output} & \textbf{Post-processed Output} \\
\midrule
1 & \emph{timi lai kasto cha sathe aile all good xau ki k xa} & \emph{timi lai kasto chha saathi aaile \textcolor{prepblue}{all good} chhau ki k chha} & \dv{तिमीलाई कस्तो छ साथी अहिले} \textcolor{rawred}{\dv{अल गुड}} \dv{छौ कि के छ} & \dv{तिमीलाई कस्तो छ साथी अहिले} \textcolor{fixgreen}{all good} \dv{छौ कि के छ} \\
\midrule
2 & \emph{ajkal kasto dherai traffic jam huna thaleko xa yll kathmandu ma} & \emph{ajkal kasto dherai \textcolor{prepblue}{traffic jam} huna thaaleko chha yaar kathmandu ma} & \dv{आजकल कस्तो धेरै} \textcolor{rawred}{\dv{ट्राफिक जाम}} \dv{हुन थालेको छ यार काठमाडौं म} & \dv{आजकल कस्तो धेरै} \textcolor{fixgreen}{traffic jam} \dv{हुन थालेको छ यार काठमाडौं म} \\
\midrule
3 & \emph{pulchowk campus ma final major project submission hudhai xw} & \emph{pulchowk \textcolor{prepblue}{campus} ma \textcolor{prepblue}{final major project submission} hudhai chha} & \dv{पुल्चोक} \textcolor{rawred}{\dv{क्याम्पसमा फाइनल मेजर प्रोजेक्ट सबमिसन}} \dv{हुँदै छ} & \dv{पुल्चोक} \textcolor{fixgreen}{campus} \dv{मा} \textcolor{fixgreen}{final major project submission} \dv{हुँदै छ} \\
\midrule
4 & \emph{literally bro yo transliterator tw as expected ramro kam gariraxa tw} & \emph{\textcolor{prepblue}{literally bro} yo \textcolor{prepblue}{transliterator} ta \textcolor{prepblue}{as expected} raamro kaam garirachha ta} & \textcolor{rawred}{\dv{लिटरेली ब्रो}} \dv{यो} \textcolor{rawred}{\dv{ट्रान्सलिटरेटर}} \dv{त} \textcolor{rawred}{\dv{एस एक्सपेक्टेड}} \dv{राम्रो काम गरिरछ त} & \textcolor{fixgreen}{literally bro} \dv{यो} \textcolor{fixgreen}{transliterator} \dv{त} \textcolor{fixgreen}{as expected} \dv{राम्रो काम गरिरछ त} \\
\bottomrule
\end{tabular}
\caption{Post-processing traces for four code-mixed inputs. \textcolor{prepblue}{Blue} = English tokens detected in the pre-processed input; \textcolor{rawred}{red} = transliterated English tokens in raw model output; \textcolor{fixgreen}{green} = correctly restored English tokens after post-processing.}
\label{tab:post-process}
\end{table}

\section{Human Evaluation Guidelines}
\label{sec:likert_guidelines}

Table~\ref{tab:likert_guidelines} lists the five-point accuracy and naturalness guidelines provided to annotators for human evaluation.

\begin{table}[H]
\centering
\small
\renewcommand{\arraystretch}{1.25}
\begin{tabular}{@{}c >{\raggedright\arraybackslash}p{6.6cm} >{\raggedright\arraybackslash}p{6.6cm}@{}}
\toprule
\textbf{Score} & \textbf{Accuracy} & \textbf{Naturalness} \\
\midrule
5 & Perfect transliteration. & Reads exactly like naturally written Nepali. \\
4 & Minor spelling variations; meaning fully preserved. & Mostly natural with only minor awkwardness. \\
3 & Noticeable errors but still understandable. & Moderately natural and readable. \\
2 & Several incorrect transliterations affecting readability. & Difficult to read naturally. \\
1 & Incorrect transliteration; intended meaning largely lost. & Unnatural or unreadable Nepali. \\
\bottomrule
\end{tabular}
\caption{Guidelines provided to annotators for human evaluation.}
\label{tab:likert_guidelines}
\end{table}

\section{Ablation Results for All Models}
\label{sec:ablations}

This appendix provides the complete ablation study results for all evaluated model configurations, detailing the impact of different preprocessing strategies (Rule-Based Normalization and PVD Normalization) on Character Error Rate (CER), Word Error Rate (WER), and their corresponding accuracies (Table~\ref{tab:full_ablations}).

\begin{table}[H]
\centering\small
\begin{tabular}{@{}lcccc@{}}
\toprule
\textbf{Configuration} & \textbf{CER} & \textbf{WER} & \textbf{\makecell{Character\\Acc.}} & \textbf{\makecell{Word\\Acc.}} \\
\midrule
\multicolumn{5}{@{}l}{\textit{Model: ByT5 Original}} \\
\quad \textbf{Rule-Based + PVD Normalization} & \textbf{6.03} & \textbf{23.23} & \textbf{93.97} & \textbf{76.77} \\
\quad PVD Normalization & 7.23 & 24.91 & 92.77 & 75.09 \\
\quad Rule-Based Normalization & 7.41 & 26.61 & 92.59 & 73.39 \\
\quad Raw (No Normalization) & 8.64 & 28.47 & 91.36 & 71.53 \\
\midrule
\multicolumn{5}{@{}l}{\textit{Model: ByT5 Augmented}} \\
\quad \textbf{Rule-Based + PVD Normalization} & \textbf{5.96} & \textbf{21.85} & \textbf{94.04} & \textbf{78.15} \\
\quad PVD Normalization & 6.46 & 22.32 & 93.54 & 77.68 \\
\quad Rule-Based Normalization & 6.65 & 23.55 & 93.35 & 76.45 \\
\quad Raw (No Normalization) & 7.24 & 24.39 & 92.76 & 75.61 \\
\midrule
\multicolumn{5}{@{}l}{\textit{Model: mT5 Original}} \\
\quad \textbf{Rule-Based + PVD Normalization} & \textbf{17.45} & \textbf{37.20} & \textbf{82.55} & \textbf{62.80} \\
\quad PVD Normalization & 19.13 & 40.08 & 80.87 & 59.92 \\
\quad Rule-Based Normalization & 19.21 & 39.58 & 80.79 & 60.42 \\
\quad Raw (No Normalization) & 21.39 & 43.24 & 78.61 & 56.76 \\
\midrule
\multicolumn{5}{@{}l}{\textit{Model: mT5 Augmented}} \\
\quad \textbf{Rule-Based + PVD Normalization} & \textbf{13.50} & \textbf{31.52} & \textbf{86.50} & \textbf{68.48} \\
\quad Rule-Based Normalization & 14.04 & 32.08 & 85.96 & 67.92 \\
\quad PVD Normalization & 14.28 & 32.96 & 85.72 & 67.04 \\
\quad Raw (No Normalization) & 18.04 & 36.83 & 81.96 & 63.17 \\
\bottomrule
\end{tabular}
\caption{Comprehensive ablation study results across all model variants and preprocessing configurations. The best configuration for each model is shown in bold.}
\label{tab:full_ablations}
\end{table}

\section{Comparison of Four Model Variants}
\label{sec:model_comparison}

To further illustrate qualitative differences among model variants, Table~\ref{tab:model_comparison} presents outputs from ByT5 Augmented, ByT5 Original, mT5 Augmented, and mT5 Original on two representative code-mixed inputs.

\begin{table}[H]
\centering
\small
\setlength{\tabcolsep}{4pt}
\renewcommand{\arraystretch}{1.3}

\begin{tabular}{@{}l p{10cm}@{}}
\toprule
\textbf{Model} & \textbf{Output} \\
\midrule
\multicolumn{2}{@{}l}{\textbf{Input 1:} \emph{timlai kasto cha sathe aile all good chau ki k chha}} \\
\cmidrule(r){1-2}
ByT5 Augmented & \dv{तिमीलाई कस्तो छ साथी आइले अल गुड छौ कि के छ} \\
ByT5 Original  & \dv{तिमीलाई कस्तो छ साथे अहिले अल गुड छौ कि क छ} \\
mT5 Augmented  & \dv{तिमीलाई कस्तो छ साठी अइले अइले हाइट छौं की खा} \\
mT5 Original   & \dv{तिमीलाई कस्तो छैन कि के छ} \\
\midrule
\multicolumn{2}{@{}l}{\textbf{Input 2:} \emph{mero desh nepal maa aba aarthik samunnati aaula}} \\
\cmidrule(r){1-2}
ByT5 Augmented & \dv{मेरो देश नेपालमा अब आर्थिक समुन्नति आउला} \\
ByT5 Original  & \dv{मेरो देश नेपाल मा अब आर्थिक समुनाति आउला} \\
mT5 Augmented  & \dv{मेरो देश नेपालमा अब आर्थिक समृद्धि आउला} \\
mT5 Original   & \dv{मेरो देश नेपालमा अब आर्थिक संकटान आउछ} \\
\bottomrule
\end{tabular}

\caption{Comparison of outputs across four model variants for two representative code-mixed inputs.}
\label{tab:model_comparison}
\end{table}

\section{Qualitative Comparison with Existing Transliteration Systems}
\label{sec:existing_systems_qualitative}

To complement the quantitative comparison in Table~\ref{tab:baselines}, we present representative examples illustrating qualitative differences between our system and existing publicly available Nepali transliteration tools (Table~\ref{tab:existing_systems_qualitative}).

\begin{table}[H]
\centering
\small
\renewcommand{\arraystretch}{1.3}
\begin{tabular}{@{}lp{9.5cm}@{}}
\toprule
\textbf{System} & \textbf{Output} \\
\midrule
\multicolumn{2}{@{}l}{\textbf{Input 1:} \emph{ma pani dherai meheant garera vlog banauxu herdinu hola}} \\
\cmidrule(r){1-2}
\textit{Reference Output} & \dv{म पनि धेरै मेहेनत गरेर भ्लग बनाउँछु हेरिदिनु होला} \\
\cmidrule(r){1-2}
ByT5 Norm. (Ours) & \dv{म पनि धेरै मेहेनत गरेर भ्लग बनाउँछु हेरिदिनु होला} \\
ByT5 Raw (Ours) & \dv{म पनि धेरै मेहेन्ट गरेर भ्लग बनाउँछु हेर्दिनु होला} \\
NepaliXlit \cite{patel2026from} & \dv{मा पनि धेरै मेहन्त गरेर भ्लग बनाउछु हेर्दिनु होला} \\
Google Transliteration \cite{googletranslit} & \dv{मा पनि धेरै मेहेंत गरेर व्लोग बनौक्षु हेर्दिनु होला} \\
Easy Nepali Typing \cite{easynepalityping} & \dv{मा पनि धेरै मेहेंत गरेर व्लोग बनौक्षु हेर्दिनु होला} \\
Ashesh Unicode \cite{asheshunicode} & \dv{म पनि धेरै मेहेअन्त गरेर व्लोग बनौक्सु हेर्दिनु होल} \\
Kokil Unicode \cite{kokilunicode} & \dv{म पानी धेरै मेहेअन्त गरेर व्लोग बनौक्सु हेर्दिनु होल} \\
Nepali Unicode \cite{nepaliunicodecom} & \dv{म पनि धेरै मेहेअन्त गरेर व्लोग बन्औक्सु हेर्दिनु होल} \\
\midrule
\multicolumn{2}{@{}l}{\textbf{Input 2:} \emph{daju le description ko lagi ai use garnu huda raixa}} \\
\cmidrule(r){1-2}
\textit{Reference Output} & \dv{दाजुले डिस्क्रिप्सनको लागि एआई युज गर्नु हुँदो रैछ} \\
\cmidrule(r){1-2}
ByT5 Norm. (Ours) & \dv{दाजुले डेस्क्रिप्सनको लागि एआई युज गर्नु हुँदै रैछ} \\
ByT5 Raw (Ours) & \dv{दाजुले डेस्क्रिप्सनको लागि एआई युज गर्नु हुँदा रहेछ} \\
NepaliXlit \cite{patel2026from} & \dv{दाजु ले डिस्क्रिप्सन को लागि आई युज गर्नु हुदा रैछ} \\
Google Transliteration \cite{googletranslit} & \dv{दाजु ले देस्च्रिप्तिओन को लागि ई उसे गर्नु हुदा रैक्ष} \\
Easy Nepali Typing \cite{easynepalityping} & \dv{दाजु ले देस्च्रिप्तिओन को लागि ई उसे गर्नु हुदा रैक्ष} \\
Ashesh Unicode \cite{asheshunicode} & \dv{दाजु ले देस्कृप्सन को लागि ऐ उसे गर्नु हुँदा रैक्स} \\
Kokil Unicode \cite{kokilunicode} & \dv{दजु ले देस्कृप्तिओन को लगी ऐ उसे गर्नु हुदा रैक्स} \\
Nepali Unicode \cite{nepaliunicodecom} & \dv{दजु ले देस्च्रिप्तिओन को लगि ऐ उसे गर्नु हुद रैक्स} \\
\bottomrule
\end{tabular}
\caption{Qualitative comparison of outputs from our system and existing Nepali transliteration tools on representative social-media inputs.}
\label{tab:existing_systems_qualitative}
\end{table}

\end{document}